\documentclass{article} % For LaTeX2e
\usepackage{iclr2021_conference,times}

\usepackage{amsmath,amsfonts,bm}

\def\eqref#1{equation~\ref{#1}}
\def\1{\bm{1}}

\DeclareMathAlphabet{\mathsfit}{\encodingdefault}{\sfdefault}{m}{sl}
\SetMathAlphabet{\mathsfit}{bold}{\encodingdefault}{\sfdefault}{bx}{n}

\usepackage{hyperref}
\usepackage{url}
\usepackage{graphicx}
\usepackage{header}
\usepackage{booktabs}
\usepackage{multirow}
\usepackage{subcaption}
\usepackage{wrapfig}
\usepackage{adjustbox}
\usepackage{enumitem}

\newcommand{\dataname}{OTT-QA\xspace}

\newcommand{\datanamefull}{{O}pen {T}able-and-{Te}xt % agg{r}egation 
Question Answering\xspace}
\newcommand{\eat}[1]{}
\newcommand{\nlp}[1]{\texttt{\small #1}}

\newcommand\cls{\texttt{[CLS]}\xspace}

\newcommand\bertsub[1]{\mathtt{BERT_{#1}}\xspace}

\title{Open Question Answering over Tables\\ and Text}

\author{Wenhu Chen$^1$
\thanks{Part of this work was done during an internship at Google.}
, Ming-Wei Chang$^2$, Eva Schlinger$^2$, William Wang$^1$, William W. Cohen$^2$\\
$^1$University of California, Santa Barbara\\
$^2$Google Research\\
\texttt{\{wenhuchen, william\}@cs.ucsb.edu}\\
\texttt{\{mingweichang, eschling, wcohen\}@google.com}
}

\iclrfinalcopy % Uncomment for camera-ready version, but NOT for submission.
\begin{document}

\maketitle

\begin{abstract}
In open question answering (QA), the answer to a question is produced by retrieving and then analyzing documents that might contain answers to the question.  Most open QA systems have considered only retrieving information from unstructured text.  Here we consider for the first time open QA over {\em both} tabular and textual data and present a new large-scale dataset \emph{Open Table-and-Text Question Answering} (OTT-QA) to evaluate performance on this task\footnote{Data was released in https://github.com/wenhuchen/OTT-QA by UCSB NLP Group}. Most questions in OTT-QA require multi-hop inference across tabular data and unstructured text, and the evidence required to answer a question can be distributed in different ways over these two types of input, making evidence retrieval challenging---our baseline model using an iterative retriever and BERT-based reader achieves an exact match score less than 10\%. We then propose two novel techniques to address the challenge of retrieving and aggregating evidence for OTT-QA. The first technique is to use ``early fusion'' to group multiple highly relevant tabular and textual units into a fused block, which provides more context for the retriever to search for.  The second technique is to use a cross-block reader to model the cross-dependency between multiple retrieved evidence with global-local sparse attention. Combining these two techniques improves the score significantly, to above 27\%.
\end{abstract}

\section{Introduction}
Open question answering considers the problem of retrieving documents from a fixed corpus with a {\em retriever}, and then analyzes retrieved evidence to provide answers to a given question with a {\em reader}. Prior open question answering systems focused only on retrieving and reading free-form passages or documents. However, a significant amount of real-world information is stored in other forms, such as semi-structured web tables due to its compact representation to aggregate related information. For example, tables are often used to hold large quantities of related facts, especially numeric facts, such as \nlp{`Career Statistics for Lebron James'}. This type of detailed information is found much less frequently in unstructured text. Tables are also commonly used for collections of homogeneous entities or recurring events, like \nlp{`List of Periodic Comets'} or \nlp{`List of Champions League Winners since 1966'}. Hence tabular information serves as an excellent complement to textual data, especially in the open setting.  Despite these advantages, no previous studies have exploited the millions of web tables to augment their open QA system.

\begin{figure}[t]
    \centering
    \includegraphics[width=0.98\linewidth]{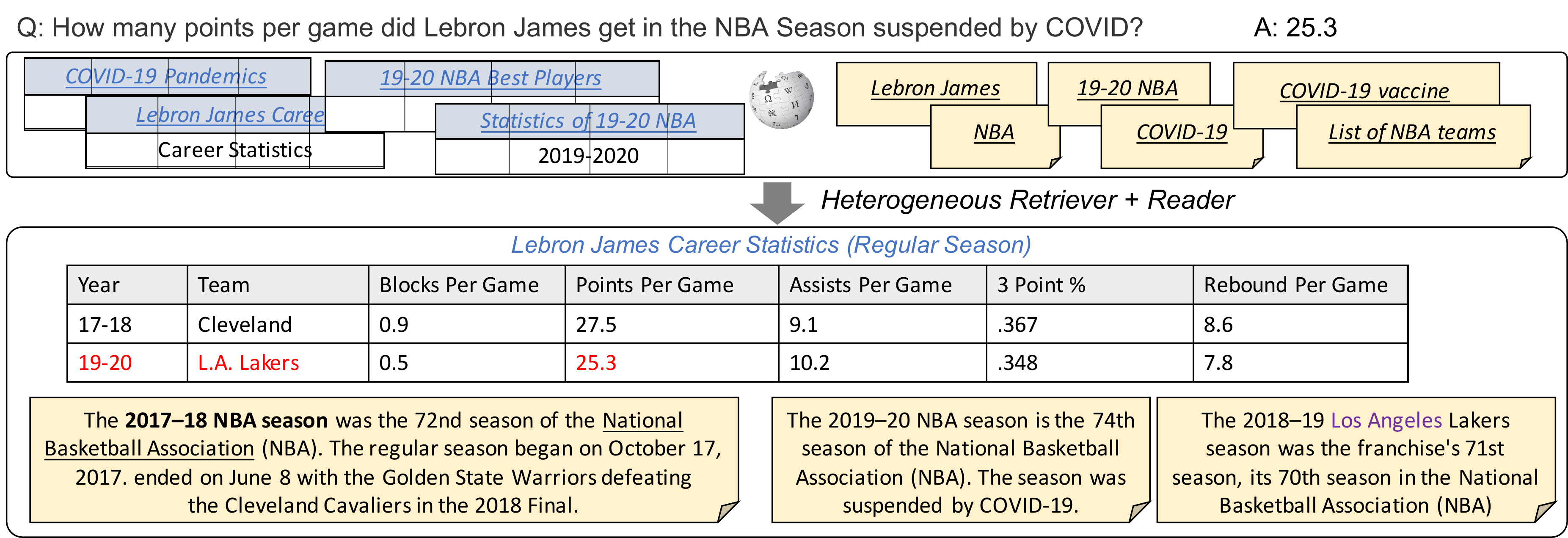}
    \vspace{-1ex}
    \caption{The problem setting:  A \dataname model needs to retrieve from two candidate pools and then perform multi-hop reasoning to find answers. }
    \label{fig:intro}
    \vspace{-3ex}
\end{figure}

In this paper, we describe the first study to jointly exploit tables and text for open-domain question answering.
For this purpose, we construct a new dataset, { \datanamefull (\dataname)}. \dataname{} is built on the HybridQA dataset \citep{chen2020hybridqa}, and like HybridQA, \dataname{} questions are multi-hop questions which require aggregating information from both tables and text to answer.  However, unlike HybridQA, \dataname{} requires the system to \emph{retrieve} relevant tables and text --- in contrast, in HybridQA, the ground truth tables and textual passages required for each question are given. To produce \dataname{}'s questions, we begin by re-annotating the questions from HybridQA to `decontextualize' them---i.e., we make questions suitable for the open-domain setting so that unique answers can be determined from the question alone, without needing context from the provided text and tables.
We then add new questions to remove potential biases. After these steps, \dataname contains 45$K$ human-annotated questions that require retrieving and aggregating information over tables and text from the whole Wikipedia. Examples from \dataname are depicted in~\autoref{fig:intro}. Note the table and passages contain non-overlapping information, and both of them must be understood to answer the question. For example, the question has a low lexical overlap with the passage about the \nlp{`Lakers'}, and it needs the table as the bridge to retrieve this passage. Such cross-modality multi-hop retrieval features \dataname. More examples are displayed in Appendix.

\dataname is distinguished from the existing QA datasets in two aspects. Existing table-based QA datasets~\citep{pasupat2015compositional,yu2018spider,chen2020hybridqa} operates in the closed setting without requiring any retrieval, whereas most existing open QA datasets ~\citep{joshi2017triviaqa, yang2018hotpotqa} require only text retrieval, not table retrieval. One dataset, Natural Questions (NQ) \citep{kwiatkowski2019natural} includes some tabular information in its corpus, but the tables are nearly always of a restricted type (infobox tables with only a single row). In contrast, \dataname models require retrieving {\em both} tabular data and text, and unlike the NQ dataset, requires information fusion from text and tables in non-trivial ways. \dataname poses novel and realistic challenges to both the retriever and reader in open QA though the questions are less natural than the real queries from NQ~\citep{kwiatkowski2019natural}. Retrievers for \dataname need to consider two information formats, making the search space larger. Even worse, as questions in \dataname often require multi-hop inference, one round of retrieval is often not enough. Readers for \dataname also need to aggregate a significant amount of knowledge-intensive information, compared to other reader models: a single table in  \dataname has an average length of over 300 words. Moreover, readers are often expected to process multiple retrieved units due to the uncertainty in retrieval, which makes it difficult to design strong reader models~\citep{devlin2019bert,liu2019roberta} with a length limit of 512 tokens.

% The plausible baseline and its shortcoming
The baseline system that we propose to address these challenges uses an iterative retriever~\citep{sun-etal-2019-pullnet,qi2019answering,min2019knowledge,ding2019cognitive,asai2019learning} and a BERT reader~\citep{devlin2019bert}. The iterative retriever explores multiple evidence documents iteratively, interacting with the candidate pool to gradually reformulate the query. Beam search is used to find multiple subsets of documents that may contain all the required evidence, and each subset is then fed to the BERT reader to predict the answer span. The highest-scored prediction is chosen as the answer. The iterative retriever needs to re-encode the query with a big transformer and re-search over the candidate pool, such a procedure (especially dense) can be computationally expensive. Furthermore, the BERT reader fails to capture a global overview of the retrieved documents, which leads to bad local optimum in the model prediction.

We propose a more sophisticated system that addresses these challenges with two novel strategies: namely {\em fusion} retrieval and {\em cross-block} reading. The {\bf fusion retriever} first pre-aligns the table segments to their highly related passages, using entity linking. Then, the aligned table segments and passages are grouped as a \emph{fused block}, which contains aggregated information from two modalities; hence, compared to the previous documents, it contains richer context to benefit the following retrieval. We view the fused block as the basic unit to be retrieved, and instead of performing multiple runs of retrieval iteratively, the fusion retriever is used once to retrieve the top $K$ fused blocks; however, due to errors in fusion and retrieval, the retrieved top-1 fused block might not contain the necessary information.  We thus also propose a {\bf cross-block reader} based on a sparse-attention based transformer architecture~\citep{ainslie2020etc,zaheer2020big}, which can process extremely long sequences efficiently.  We use the cross-block reader to read all the top-K retrieved fused blocks jointly. Both strategies have proven effective compared to the baseline system: the best model combining the two strategies improves the accuracy of the baseline system by a huge margin. 
\section{Background}
%maybe say what is open and closed here
The aim of an open QA system is to extract an answer to a question $q$ from a given large corpus. Most open QA models are retriever-reader models, which extract answers in two steps: retrieval and reading. In the {\em retrieval} step, a retrieval model $f$ is used to retrieve a set of passages from the text corpus. In the {\em reading} step, the reader is then used to extract the answer from them.
\vspace{-2ex}
\paragraph{Retrieval Function}
\label{sec:retrieval}
There are two commonly-used types of retrieval function $f$: sparse retrievers and dense retrievers. Our sparse retriever uses a unigram-based BM-25 score to retrieve an evidence unit $b$ from the candidate pool $\mathbb{B}$. Our dense retrieval function is a dual-encoder model~\citep{bromley1994signature}, and we follow \citep{lee2019latent,guu2020realm} for the dual encoder design. The query and the passage are encoded with separate Transformers. As in \citep{devlin2019bert}, the vector corresponding to the first token, $\cls$, is used as a ``pooled'' representation of the sequence. The dense retrieval function is the dot product between $h_q=\bertsub{Q}(q)$[CLS] and $h_b=\bertsub{B}(b)$[CLS] for each evidence block $b$ in the candidate corpus---i.e., the scoring function is $f(q, b) = h_q^T h_b$, which can viewed as finding the nearest neighbor in vector space. In the multi-hop open QA setting~\citep{yang2018hotpotqa}, an iterative retrieval function~\citep{sun-etal-2019-pullnet,min2019knowledge,ding2019cognitive} is proposed, which defines the retrieval process as an auto-regressive formula. Our iterative retriever function is denoted as $f([q, b_1, \cdots, b_{j-1}], b_j)$, which appends the previous $j-1$ rounds of retrieval to the original $q$ in in the $j$-th round of retrieval. Beam search is used in test time.
\vspace{-2ex}
\paragraph{Single-Block Reader}
Due to the uncertainty in retrieval, the top-1 document might not always contain the answer. Existing models normally retrieve the top-$k$ documents and feed them to the reader for span selection. The standard reader~\citep{chen2017reading,joshi2017triviaqa} aims to extract a span from each of the retrieved blocks $b_i$ and assign a confidence $f(q, b_i)f_{read}(a |q, b_i)$ to it, with $f(q, b_i)$ indicating the retrieval probability and $f_{read}(a |q, b_i)$ denoting the span selection probability by reader. Multiple answers $\{a_1, \cdots, a_k\}$ are ranked with this confidence score and the highest scored answer span $\hat{a}$ is the final answer.
Note that the reader needs to run $k$ times, once for each of the top-$k$ retrievals.  We refer to this model as the {\em single-block reader} and use it as our baseline.
\vspace{-2ex}
\paragraph{HybridQA}
HybridQA~\citep{chen2020hybridqa}, a closed-domain QA dataset, is the most related to ours. During the annotation of HybridQA, a table $T$ and its relevant passages $\{P_1, \cdots, P_N\}$ (surrounding text and hyperlinked passage) are given to a crowd worker to write questions which necessarily require both the passage and table information to answer. The original dataset contains 72K multi-hop questions paired with 13K tables with their paired passages. During training/testing time, the ground-truth tables and passages are given to a model, HYBRIDER, to find the final answer. HYBRIDER also serves as an important baseline in our paper. %Since we do not have `ground-truth' input, we use the retrieved table and passage as input.
\section{Task and Dataset}
In \dataname, the retrieval corpus consists of a set of table candidates $\mathbb{B}_T$ and a set of passage candidates $\mathbb{B}_P$. The task is to answer question $q$ by extracting answer strings from blocks $b \in \mathbb{B}_T \cup \mathbb{B}_P$, where $b$ can be either textual and tabular data.  We adopt the standard exact match (EM) and F1 scores~\citep{yang2018hotpotqa} for evaluation. Different from HybridQA, \dataname{}'s table candidates are web tables \emph{without} hyperlinks provided. This decision was made to make the problem setting more general, as otherwise systems that solve \dataname{} could only be applied to high-quality data in Wikipedia. However, in \dataname, we provide hyperlinks in the training subset, but not dev/test set. Removing hyperlinks in tables makes the overall task much more challenging, but makes the final systems applicable to more general domains.  Thus, an \dataname model needs to jointly retrieve both tables and text, without abusing gold hyperlinks, and then aggregate them to find the answer. 
\vspace{-2ex}
\paragraph{Candidate Pool}
For our table collection $\mathbb{B}_T$, we extracted all Wikipedia regular tables with their metadata including page title, page section title, and section text. The metadata, denoted $T_M$, is essential for de-contextualization. We obtain a table corpus containing over 400$k$ high-quality tables with an average length of 320 words including metadata. For the text passage collection  $\mathbb{B}_P$, we crawl English Wikipedia dump pages and filter out noisy pages. We follow HybridQA~\citep{chen2020hybridqa} and only keep a maximum of 12 sentences in the introduction section as the passage. We obtain a corpus containing over 5 million passages, with an average of 94 words.
\vspace{-2ex}
\paragraph{Notation}
We define each table as a matrix $T$, which consists of cells $T_{i,j}$ with $i$ specifying the row, and $j$ specifying the column. Each cell $T_{i,j}$ could be a number, date, phrase or even sentence due to its semi-structured nature. However, a single complete table with structured representation~\citep{herzig2020tapas} can easily exceed the 512-token limit, which poses great challenges to the downstream reader to process top-$K$ retrieval. Hence we propose to decompose each table $T$ into multiple rows $R_i$, which are combined with the headers, metadata, and global max/min information from the original table as a {\bf table segment}. The table segment is used as the basic retrieval block in our paper. This decomposition procedure increases candidate $\mathbb{B}_T$ from 400$k$ to 5 million, making the retrieval problems even more fine-grained and more challenging. Our table segment representation is described in Appendix~\autoref {section:table_seg_representation}. In summary, we build a candidate pool of 5 million table segments $\mathbb{B}_T$ and a pool of 5 million passages $\mathbb{B}_P$. We denote as $\mathbb{B}$ as our {full candidate pool}, which our model needs to find the block $b$ (a table segment or a passage) containing the answer span.

\subsection{Question and Answer Annotations}
\begin{figure}[!thb]
    \centering
    \includegraphics[width=0.98\linewidth]{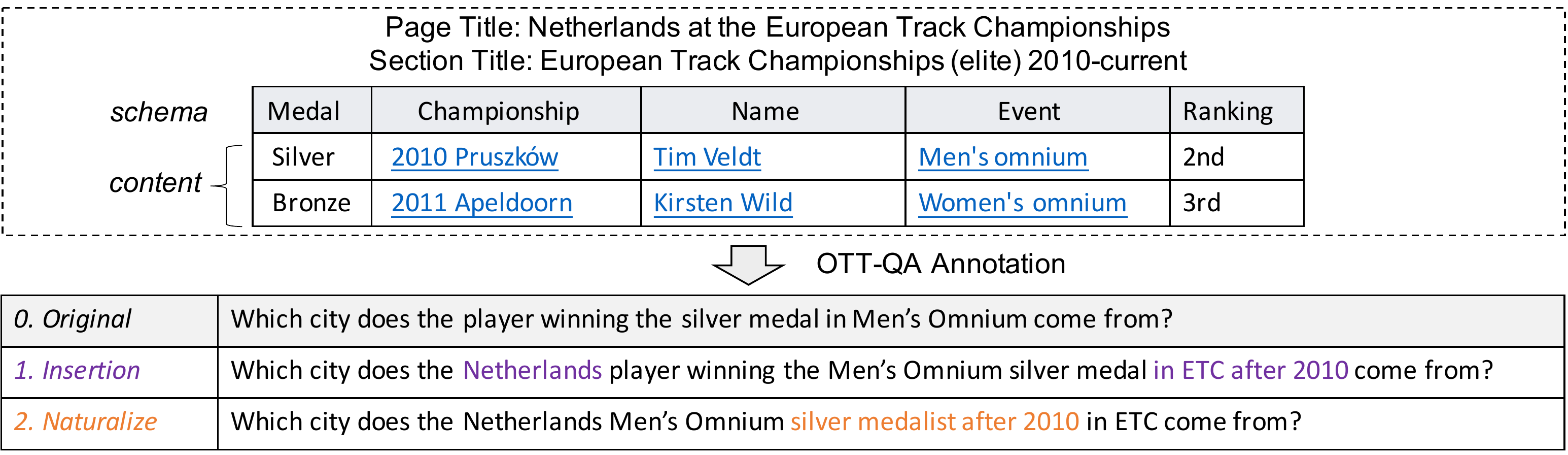}
    \caption{The `de-contextualization' annotation phase of \dataname. In the first step, the annotator is restricted to add phrases from the context. In the second step, the annotator is specifically requested to make the sentence more concise and natural. }
    \label{fig:instruct}
    \vspace{-2ex}
\end{figure}

Our question and answer pairs are built upon the existing HybridQA~\citep{chen2020hybridqa} dataset, with several significant changes. First, crowd workers `decontextualize' the questions so that they are not under-specified or context-dependent, and thus suitable for the open setting. Second, we add more questions to the development/test set to remove possible annotation bias. During annotation, we adopt strict quality control\footnote{The collection is conducted on an established crowdsourcing platform with annotators from countries with English as the native language. The annotators were required to meet the requirements: 1) a native speaker in an English-speaking country 2) having an approval rate of over 95\% and 3) having at least 500 approved jobs.}
 and more details are described in Appendix~\autoref{sec:quality_control}.
\vspace{-2ex}
\paragraph{Decontextualization}
Most questions in HybridQA are contextualized with
a given table and several passages, with corresponding questions written by crowd workers. Often, the crowd-sourced questions assume the context. For example, the questions might contain the words \nlp{"the players"} because the given table is about \nlp{"Netherlands players"}. We thus needed `de-contextualize'~\citep{parikh2020totto} the original context-dependent questions, so they could serve as standalone questions, specific enough to imply a unique answer relative to the corpus. To discourage excessive unwanted modification, we enforce a two-step annotation procedure, as depicted in~\autoref{fig:instruct}. In the first phase, the worker is only allowed to insert {\bf minimum} words or phrases (or replace pronouns) into the questions based on the information presented by Wikipedia Title, Section Title, and Section Text to make the question have a unique answer. After this step, we often potentially obtain overly-complicated questions that are artificial and unnatural. Therefore, we manually selected the worst 25\% questions and sent them back to make them more concise and natural.

\paragraph{Additional Evaluation Examples}
As all the questions from HybridQA are based on the $13k$ tables from the HybridQA set, no questions are asked about the newly crawled 400$k$ tables.  This potentially generates unwanted statistical biases or artifacts for the model to exploit, and potentially biases the final evaluation results. Therefore, we randomly sampled another 1100 tables from the newly crawled tables, and follow the original annotation process used by HybridQA to re-collect 2200 new questions.  These new questions were mainly used in the dev/test set. Below we refer to the subset of tables used by original HybridQA as the {\em in-domain} tables.
\vspace{-2ex}
\paragraph{Distant Supervision Signals} 
For the in-domain tables ($\approx$ 8$k$), the cell-wise hyperlinks are provided in \dataname{} as a potential signal for supervision. We use $H_{i,j} = \{b_1, b_2, \ldots \in \mathbb{B}_P \}$ to denote the hyperlinks in cell $T_{i,j}$.  Since in HybridQA the oracle fine-grained answer span is not explicitly annotated, we approximate this by traversing the table and hyperlinked pasasages to find all exact matches. This process contains some noise---a manual study reveals that it roughly contains 15\% error. We use this `weakly-supervised' fine-grained information to train our models. We denote the `approximate' block of the answer span for answer $a$ as $b_a$, and use it to train our model.

\subsection{Dataset Statistics}
After annotation, we sampled roughly 2K questions from the in-domain HybridQA dataset, and then mix them with the newly collected out-domain questions to construct our dev and test sets. Finally, we have 41,469 questions in the training set, 2,214 questions in the dev set, and 2,158 questions in the test set. We conduct more in-detailed analysis over the reasoning types and show them in the Appendix~\ref{section:question_type}, a remarkable difference from original HybridQA is that a proportion of questions actually have multiple plausible inference chains in the open-domain setting.

%The statistics of \dataname is described in~\autoref{tab:statistics}.
%\input{figures_tables/dataset_stats}

\section{Model}

Our model for \dataname is a retriever-reader model with new designs for both retriever and reader. As discussed briefly above, we propose to use a fusion retriever instead of using a standard iterative retrieve, and we also propose to use cross-block readers to replace a standard single-block reader.

\begin{figure}[!thb]
    \centering
    \includegraphics[width=0.95\linewidth]{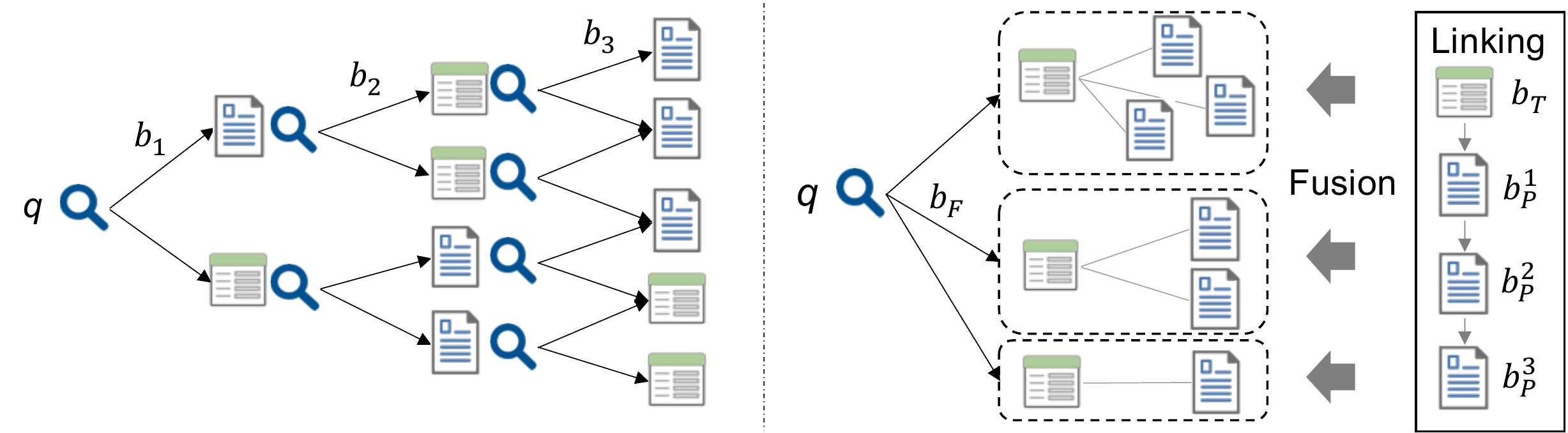}
    \vspace{-1ex}
    \caption{{\bf Left:} Iterative 3-step retrieval over individual blocks (baseline). {\bf Right:} Fusion 1-step retrieval over fused groups, which greatly lowers the cost of iterative encoding and retrieving. }
    \vspace{-2ex}
    \label{fig:retriever}
\end{figure}

\subsection{Fusion Retriever}
Iterative retrieval (Figure~\ref{fig:retriever}, Left)
has the following issues. First, iterative retrieval training often requires having supervision signals for every retrieval step to reach good performance, which is not available in \dataname.
The iterative retrieval also suffers
from the problem of error propagation, as early mistakes can propagate to later retrieval stages. Finally, the computation cost for applying a dual-encoder for iterative retrieval is very high, as for every stage, the query embedding has to be re-encoded to include the entire retrieval history. 

We propose an alternative strategy to replace multi-step retrieval, namely fusion retrieval (Figure~\ref{fig:retriever}, Right). In the fusion retriever, we first use an `early fusion' strategy to group relevant heterogeneous data before retrieval. The fusion procedure groups several highly-relevant blocks from different modalities as a self-contained group (fused block), which provides more clues for the retriever to utilize. Early fusion is very important for retrieving table segments, which often have incomplete context by themselves. The early fusion process aims to fuse a table segment and relevant passages into a group. Here we propose to fuse entities mentioned in a table segment to the appropriate passages for those entities; this is similar to document expansion based on a traditional entity linking step. The problem is challenging due to the mismatch between the lexical forms from the table (which for brevity are often abbreviated) and the relevant passage titles. For example, a cell in the table of \nlp{"NCAA Division I Men's Football Tournament"} contains the term \nlp{"Penn State"}. Directly matching \nlp{"Penn State"} against the passage corpus will lead to \nlp{"Penn State University"} rather than the ground-truth hyperlinked entity, named \nlp{"Penn State Nittany Lions football"}. Therefore, we propose an additional augmentation step, which takes in a table segment block $b_T$ and generates a sequence of augmented queries $q_1, q_2, \cdots, q_n$ token by token to make the queries more similar to the passage title. The augmented queries are then used to search for nearest neighbors in the passage corpus $\mathbb{B}_P$ using BM25 as the final entity linking step, which is depicted in Appendix. The query augmentation is implemented with a GPT-2 model~\citep{radford2019language}, fine-tuned on the supervised pairs of (table segment, hyperlink) from the in-domain tables. Each $b_T$ is fed to find its companions $b_P^1, \cdots, b_P^n$, they are collectively called $b_F$.

We follow the standard dual-encoder setting (\autoref{sec:retrieval}) and the only difference is that we replace the input of the block encoder with $h_b = \bertsub{B}([b_T, b_P^1, \cdots, b_P^n]).$, which captures the cross-attention between the table and the text within a block. The fused embedding contains richer context from both modalities to complement each other. The retriever only needs to retrieve once from the candidate pool, which dramatically decreases the complexity compared to the existing iterative retrievers.

To enhance the neural retrieval system to retrieve fused blocks, we apply the Inverse Cloze Task (ICT)~\citep{lee2019latent} pretraining task on the corpus of fused blocks. ICT is a way to generate pseudo-training data for dense retrieval. Unlike standard document-wise ICT, our fused block contains both table segments and multiple passages. Given a fused block $b_F$, we generate the pseudo-query in the following way: 1) we first corrupt the table segment by randomly dropping half of the words from the table metadata and cells to obtain a partial table segment $\hat{b}_T$. 2) We then randomly sample a sentence $\hat{b}_P$ from the fused passage. We combine $\hat{b}_T$ and $\hat{b}_P$ as a pseudo query $\hat{q}$ and pair it with the original fused block $b_F$ as pre-training data. The pre-training data is applied to enhance the dual encoder's ability to select lexically matched documents. After pre-training, the retriever is fine-tuned on \dataname. Finally, at inference time, the retriever is used to retrieve the top $K$ fused blocks for a question, which are then passed to the reader for answer prediction.

\subsection{Cross-Block Reader}
\begin{figure}[t]
    \centering
    \includegraphics[width=1.0\linewidth]{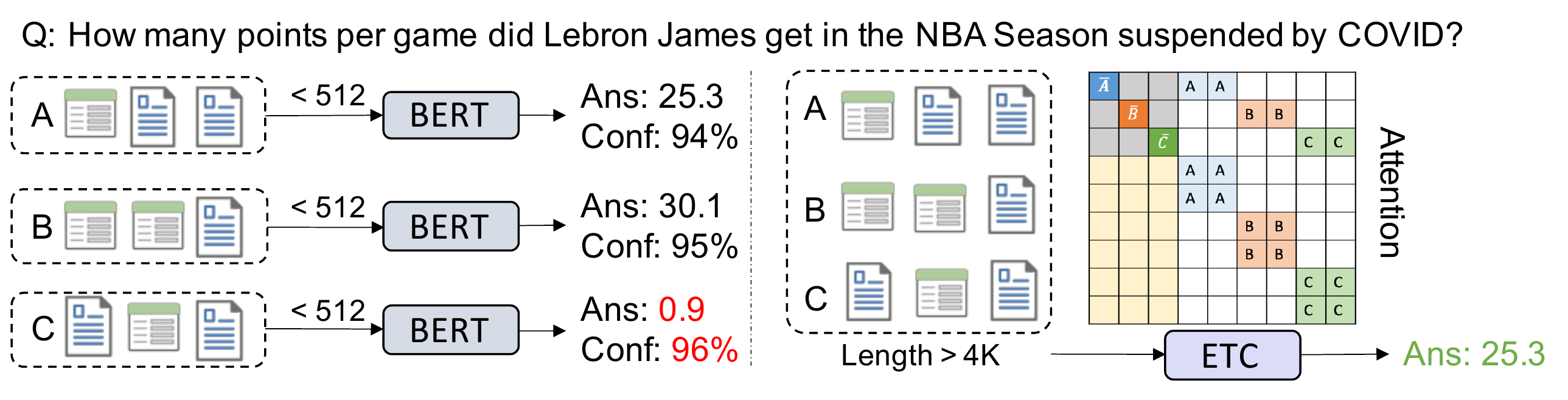}
    \vspace{-2ex}
    \caption{{\bf Left:} Single-block reader with input shorter than 512 tokens (baseline). {\bf Right:} Cross-block reader with length over 4K tokens, and $\bar{A}$ denotes the global state assigned to local block A. The single-block reader is stuck at local optimum, while cross-block reader outputs global optimum. }
    \vspace{-2ex}
    \label{fig:reader}
\end{figure}

The reader typically needs to process the top-$k$ retrieved blocks returned by the retriever to extract the best answer, as the top-1 block might not contain enough evidence to answer the question.  As demonstrated in~\autoref{fig:reader}, the cross-block reader aims to address this issue by using cross attention between different blocks to model their dependencies. To obtain the cross-block reader, we take the pre-trained long-range sparse attention transformer (ETC)~\citep{ainslie2020etc}, which can accept up to 4096 tokens as input, and then fine-tune the model on the distant supervision data. During training, the ground truth (fused) blocks are mixed with hard negative blocks from the retriever. We take the top-$k$ retrieval results to fill the 4096 token space (roughly 15 fused blocks).

Cross-attention between blocks allows a much more powerful way to aggregate information across the $k$ retrieved blocks compared to the single-block reader, especially when the blocks are fused. This is feasible because of the design of the sparse attention structure in ETC, which can constrain the attention of each token to its neighboring tokens within a local radius in its local block. Such sparse attention can decrease the attention computation complexity from quadratic $\mathcal{O}(N^2)$ to linear $\mathcal{O}(N|R|)$, where $|R|$ is the local radius (where $N=4096$ and $|R|=84$ in our experiments). To allow cross-block interaction, ETC assigns a global state for each local block in the long sequence, and blocks can attention to each other through multiple layers of such global-local structures. 
\section{Experiments}

All of our code is based on Tensorflow~\citep{abadi2016tensorflow}. For the retriever part, the sparse retriever is built on top of DrQA~\citep{chen2017reading} with unigram features, and the dense retriever is built with BERT. The single-block retriever is based on BERT-uncased, and the cross-block reader is based on ETC~\citep{ainslie2020etc}. Both of them consist of 12 layers with a hidden size of 768, the minor differences in the relative positional embedding used in ETC. All the models are trained with a learning rate of 1e-5 optimized by AdamW~\citep{loshchilov2017decoupled}. We use in-batch negatives~\citep{lee2019latent} to train our dense retrievers. A more detailed implementation of the baseline iterative retriever is described in Appendix~\autoref{section:iterative_retriever}. In fusion retriever, we use the `fused' block containing the `approximate' answer block $b_a$ as the positive instance. In iterative retriever, since the auto-regressive model $f(b_j|q,b_1,\cdots,b_{j-1})$ requires fine-grained inference chain for step-wise supervision, which is not given in \dataname. We apply lexical match based heuristics to synthesize inference chains as weakly supervised training data (described in Appendix). For all the dense retrievers, we pre-train with 10K steps using the generated pseudo query and then fine-tune them another 10K step using a batch size of 2048. For the cross-block reader, we fine-tune with a batch size of 64. Both are using 16 cloud TPUs.

\paragraph{Main Results}

\begin{table}[t]
\small
\centering
    \begin{tabular}{lcccc|cc}
        \toprule
        Retriever               & \multicolumn{2}{c}{Dev-Sparse} & \multicolumn{2}{c}{Dev-Dense} & \multicolumn{2}{|c}{Test-Best} \\
        \midrule
        Model                                     & EM     & F1    &  EM    & F1 & EM & F1    \\
        \midrule
        HYBRIDER (Top-1)~\citep{chen2020hybridqa}         & 8.7    & 10.9  & 8.9   & 11.3 &  8.4 & 10.6    \\
        HYBRIDER (best Top-K)~\citep{chen2020hybridqa}         & 9.9    & 12.2  & 10.3   & 13.0 &  9.7 & 12.8    \\
        Iterative-Retrieval + Single-Block Reader & 9.8   & 13.3    & 7.9   & 11.1  & 9.6  & 13.1 \\
        Fusion-Retrieval + Single-Block Reader   & 14.3   & 17.8  &  13.8   & 17.2  & 13.4 & 16.9  \\
        Iterative-Retrieval + Cross-Block Reader  & 17.1  & 20.7    & 14.4   & 18.5   & 16.9 & 20.9  \\
        Fusion-Retrieval + Cross-Block  Reader   & 27.7   & 31.8 & {\bf 28.1}  & {\bf 32.5} & {\bf 27.2} & {\bf 31.5} \\
        \midrule
        $\dagger$ Table-only Retrieval + Cross-Block Reader  & 4.6   & 6.9  & 4.9  & 7.2  &  4.4 & 7.0 \\
        $\dagger$ Text-only Retrieval + Cross-Block Reader  & 8.2   & 12.4  & 8.9  & 12.8  &  8.8  & 12.1  \\
        $\dagger$ Oracle Link + Fusion-Retrieval + Cross-Block  Reader & 35.8  & 40.1  & 35.2  & 39.9  & 35.0  & 39.5 \\
        $\dagger$ Oracle Table + Link (w/o Retrieval) + HYBRIDER  &  44.1 & 50.8 & 44.1 & 50.8 & 43.0 & 49.8 \\
        \bottomrule
    \end{tabular}
    \vspace{-1ex}
    \caption{{\bf Main Results}. We conduct experiments with both sparse and dense retrievers using the dev set, and then select the best setting to report the test set results (as indicated by the word "Best"). Fusion-Retriever and Cross-Block Reader are combined to obtain the highest score. $\dagger$ are ablations. }
    \label{tab:dev-test}
    \vspace{-3ex}
\end{table}
In our experiments, we experiment with different types of retriever and reader models under both sparse and dense setting, the details are described as follows:\vspace{1ex}\\
\noindent $\bullet$ HYBRIDER: this model, designed for closed domain HybridQA questions, is one baseline. Since this model requires a ground truth table with its hyperlinks to do modularized reasoning, we use BM25 to retrieve the most relevant table and passages to reconstruct an `approximated' input for this model. We experiment with top-1,2,3,4 cases where we use the answer with the highest confidence as the final result. We also directly feed the ground-truth table and hyperlinks to HYBRIDER, which roughly estimates an upper limit of this task.\vspace{1ex}\\
\noindent $\bullet$ Iterative-Retriever (Sparse):  We use a 2-step iterative retriever: in the first step, we apply the question to retrieve the top-10 table segments and top-10 passages. In the second step, we use each retrieved table segment to retrieve its related top-5 passages and concatenate each retrieved passage title with the original question to retrieve the top-5 table segments. We merge and calculate the retrieval score of each unique block and rank them by their score. For the single-block reader, we split the retrieved blocks into 512-token chunks and feed them to the BERT reader. For the cross-block reader, we truncate the top 4096 subword tokens and only feed these tokens to reader.\vspace{1ex}\\
\noindent $\bullet$ Iterative-Retriever (Dense): We use a 3-step iterative retriever. In the first step, we encode the question and retrieve the top-8 blocks (either table segment or passage); in the second step, we concatenate the previous retrieved block and the question to re-encode the query vector to further retrieve top-4 blocks; similarly, the last step retrieves top-2 blocks.\vspace{1ex}\\
\noindent $\bullet$ Fusion Retriever (Iterative): We use a sparse retriever to directly retrieve the top-15 fused blocks based on bag-of-words BM25 score, and then split it into individual table segments and passage blocks. Since passage could be associated with multiple fused blocks, we merge duplicate blocks and use their summed score. Finally, we rank each block based on its merged retrieval score and truncate the first 4096 subword tokens for the next step.\vspace{1ex}\\
\noindent $\bullet$ Fusion Retriever (Iterative): We use a dual-encoder dense retriever to directly retrieve the top-15 fused blocks, and then follow the same procedure as above. Without specifying the dense retriever uses ICT for pre-training by default.\vspace{1ex}\\
\noindent $\bullet$ Fusion Retriever w/o ICT and w/o GPT-2: these two ablation studies are aimed to show the effectiveness of our proposed ICT pre-training and query augmentation.

The main results are presented in~\autoref{tab:dev-test}. First, we can observe that best HYBRIDER top-2 can only achieve a comprised exact match of 9.9\% while the oracle HYBRIDER can obtain a score of 44\%, which reflects the difficulties of the hybrid retrieval in our dataset. We restrain the retriever to only retrieve table and text to answer the questions and report their results in~\autoref{tab:dev-test}, even with the strong cross-block reader, the model only obtains 10\% EM. These experiments demonstrate the necessity to integrate information from both forms in \dataname. 

By combining the standard iterative retriever and single-block reader, we can slightly improve the score can to roughly 10\%. By replacing the iterative retriever with the proposed sparse fusion retriever, the EM score can reach 14\%, a 4.5\% absolute improvement. By replacing the single-block with the proposed cross-block reader, the EM score can reach 17\% , a 7\% absolute improvement. However, by combining the two strategies, the final EM score can reach 28\%, with an 18\% absolute improvement, which is greater than the sum of individual improvements. The observation suggests the two components can affect each other in a positive way. We conjecture that the fusion retriever is more likely to retrieve mutually-supportive blocks in a group, which makes the multi-hop reasoning across different blocks easier for the following cross-block reader. In comparison, the iterative retriever retrieves isolated table segments and passages separately, which can easily miss out on the bridging evidence for building the complete reasoning chain. Thus, the cross-block reader cannot maximize its advantage in reasoning across blocks. 

%We found that  Fusion-Retrieval + Cross-Block  Reader (w/o ICT)  & -   & -  & 24.6  & 27.9 &  23.5 & 27.3 \\
%$\dagger$ Fusion-Retrieval + Cross-Block  Reader (w/o GPT-2)  &  21.3  &  24.8 &  22.1  & 25.3  & 21.2  & 24.7  \\

By removing the ICT pre-training and query augmentation, we observe that the Dev-EM score drops to 24.6\%. By removing the GPT-2 query augmentation, the Dev-EM performance drops to 22.1\%. These two results indicate the effectiveness of the proposed two strategies. By replacing the predicted hyperlinks with the oracle links, the fusion model performance can increase by 7\% EM. This indicates that there is still plenty of room to improve for the table-passage fusion model.

\paragraph{Linker/Retriever Results}
To understand the results more, we evaluate the standalone table-passage entity linking accuracy and retriever recall.

\begin{figure}[!thb]
\centering
\begin{minipage}[t]{.50\textwidth}
  \centering
  \begin{tikzpicture}
\begin{axis}[
    height=1.2in, 
    width=0.88\linewidth,
    scale only axis,
    ybar=3pt,
    enlargelimits=0.15,
    legend style={
     at={(0.05,0.95)},
      anchor=north west,
      legend columns=-1
      },
    ymin=30,
    ymax=60,
    xticklabel style = {font=\small,yshift=0.5ex},
    symbolic x coords={
    precision,
    recall,
    F1,
    },
    xtick=data,
    nodes near coords,
    nodes near coords align={vertical},
    every node near coord/.append style={font=\tiny},
    ]
\addplot[orange!20!black,fill=orange!70!white] coordinates {
(precision, 30.3) (recall, 44.2) (F1, 35.9)
};
\addplot[blue!20!black,fill=blue!70!white]  coordinates {
(precision, 42.7) (recall, 44.1) (F1, 43.3)
};
\addplot[yellow!20!black,fill=yellow!70!white]  coordinates {
(precision, 50.7) (recall, 50.1) (F1, 50.4)
};

\legend{BM25,Dual-Enc,GPT2}
\end{axis}
\end{tikzpicture}
  \vspace{-4ex}
  \caption{Entity linker performance (F1). }
  \label{fig:entity_score}
\end{minipage}%
\begin{minipage}[t]{.50\textwidth}
  \centering
  \begin{tikzpicture}
\begin{axis}[
    height=1.2in, 
    width=0.9\linewidth,
    scale only axis,
    ybar=2pt,
    enlargelimits=0.15,
    legend style={
     at={(0.05,0.95)},
      anchor=north west
      ,legend columns=-1
      },
    ymin=20,
    ymax=55,
    xticklabel style = {font=\small,yshift=0.5ex},
    symbolic x coords={Iter-1,
    Iter-2,
    Iter-3, 
    Fusion,
%   Fusion-Retriever (GPT)
    },
    xtick=data,
    nodes near coords,
    nodes near coords align={vertical},
    every node near coord/.append style={font=\tiny},
    ]
\addplot
[orange!20!black,fill=orange!70!white] coordinates {(Iter-1, 24.2) (Iter-2,37.2) (Iter-3, 35.8) 
(Fusion, 48.1)};

\addplot
[blue!20!black,fill=blue!70!white]  coordinates {(Iter-1, 22.8) (Iter-2,26.2) (Iter-3, 27.2) 
(Fusion, 52.4)};

\legend{sparse, dense, dense-ICT}
\end{axis}
\end{tikzpicture}
  \vspace{-4ex}
  \caption{Retriever performance (HITS@4K). }
  \label{fig:retriever_score}
\end{minipage}
\vspace{-1ex}
\end{figure}

We consider the following linking models: a) BM25 model, which directly uses the cell value to retrieve passages based on their titles without query augmentation, b) a Dual-Encoder model, which encodes the cell value and meta information into a query vector to compute dot-product over all the passage candidate to retrieve, c) a GPT-2 model, which first augments the cell value by the context and then uses BM25. We demonstrate our findings in~\autoref{fig:entity_score}, and evaluate with table-segment-wise F1 score. We observe that directly using BM25 leads to compromised precision of 30.3\%, which is mainly due to the lack of context information. By using a dual-encoder retriever, the precision can be improved to 42\%. However, many table segments have either zero or multiple linked passages and can be better modeled by an auto-regressive retrieval process.

We use HITS@4K is used to measure the retriever performance, which indicates the chance of ground truth block existing in the retrieved 4096 subword tokens. The results are reported in~\autoref{fig:retriever_score}. We vary the steps of iterative retrievers to show the necessity of multi-hop retrieval in \dataname. We observe that the 1-step retrieval has the lowest recall because the answer block in \dataname normally has a lower lexical overlap with the query. Adding the second retrieval step can greatly improve the recall, but adding the third retrieval hop has very little impact. In contrast to the iterative retriever, the fusion retriever can consistently improve the performance over the iterative setting for both sparse and dense setting. The sparse setting can rise from 35.8\% to 48.1\% indicating the advantage of `early' fusion. The dense retriever's improvement is more dramatic (from 27.2\% to 52.4\%). We believe this is because the iterative retriever heavily relies on noisy synthetic inference chain data, while the fusion retriever does not require such a fine-grained supervision signal, thus less prone to noise. To better understand the retriever, we conduct detailed error analysis in~\autoref{section:error_analysis}.

%We also observe that the single block reader cannot handle the multiple retrieval units well. When we increase the retrieved tokens from 512 $\rightarrow$ 1024 $\rightarrow$ 4096 and chunk them to multiple blocks for the reader, the final EM score drops from 13.8 $\rightarrow$ 10.7 $\rightarrow$ 7.8. We conjecture that the local soft-max normalization inside the local block makes the reader overly confident about its prediction and cause exposure bias issue in the prediction.

\paragraph{Generalization Results}
To further demonstrate our model's effectiveness on more general open-domain question answering, we also test on purely text-based multi-hop question answering dataset HotpotQA~\citep{yang2018hotpotqa}. In this dataset, we follow the same procedure to group passage individual blocks as grouped blocks offline. The retrieved blocks are fed to the cross-block readers to allow interaction between retrieved blocks. We demonstrate our results on the dev-set in~\autoref{tab:hotpotqa-dev}, where we compare against the other models using similar model size (BERT-base). As can be seen, our model is able to achieve competitive performance with these state-of-the-art models. However, our model is more efficient in a sense that our model does not require any additional re-ranking step over the retrieved blocks. Both Transformer-XH~\citep{zhao2019transformer} and RNN-Retrieval~\citep{asai2019learning} require expensive re-ranking over a large amount of retrieved reasoning graphs to select the most salient documents for the next stage (reader). In contrast, our model directly feeds the retrieved blocks to reader, which greatly simplifies and accelerates the system.

\begin{table}[t]
\small
\centering
    \begin{tabular}{lcc}
        \toprule
        Model                                     & Dev-EM     & Dev-F1   \\
        \midrule
        Semantic Retriever~\citep{nie2019revealing}         & 46.5    & 58.8  \\
        Cognitive Graph~\citep{ding2019cognitive} & 37.6   & 49.4    \\
        DocKIT~\citep{dhingra2019differentiable}   & 42.1   &  51.7  \\
        Transformer-XH~\citep{zhao2019transformer}  &  50.2 & 62.4   \\
        RNN-Retrieval~\citep{asai2019learning} (BERT base) &  52.7 & 65.8 \\
        \midrule
        Ours (Fusion Retriever + Cross-Block Reader)   & 50.4   & 61.7 \\
        \bottomrule
    \end{tabular}
    \vspace{-1ex}
    \caption{{\bf HotpotQA Results}. We conduct experiments on dev-set of HotpotQA, and then compare with the state-of-the art models with similar model size. }
    \label{tab:hotpotqa-dev}
    \vspace{-3ex}
\end{table}
\section{Related Work}

\noindent{\bf Table Retrieval:}
Tables are pervasive on the Web, there have been some studies on mining web tables to answer open-domain questions~\citep{sun2016table,chakrabarti2020open}. In ~\cite{sun2016table}, the authors have proposed a pipeline framework to first detect the topic entity and then generate a candidate chain, finally ranking chains to predict the answer cell. In ~\cite{chakrabarti2020open}, the authors investigate different similarity matching features to retrieve tables from the web. Our paper is significantly different from these two studies in two aspects: 1) the previous papers use private small-scale datasets while we collect a large-scale dataset and release it for public use, 2) the previous studies are restricted to only using tables as evidence, while our paper considers a more realistic and challenging setting with both table and text corpus. Tables have been a ubiquitous information representation form to express semi-structured information. There has been a long-standing effort to utilize tables in natural language processing applications~\citep{pasupat2015compositional,zhong2017seq2sql,yu2018spider,parikh2020totto,chen2019tabfact}. However, these existing tasks are restricted to in-domain cases without requiring any retrieval, and our paper is the first to investigate retrieving web tables for downstream tasks. Another pair of related works are TAPAS~\citep{herzig2020tapas} and TABERT~\citep{yin2020tabert}, which investigate joint pre-training over textual and tabular data. Our method draws inspiration from these models, and also uses special tokens and embeddings to encode spatial and logical operations inside tables. \vspace{1ex}\\
\noindent{\bf Long Range Transformer:}
Recently, many transformer variants to resolve the $\mathcal{O}(n^2)$ attention cost have been proposed including Sparse Attention~\citep{child2019generating}, Reformer~\citep{kitaev2020reformer}, Routing Transformer~\citep{roy2020efficient}, Longformer~\citep{beltagy2020longformer} and ETC~\citep{ainslie2020etc}. These different transformer models apply hierarchical architecture, local-sensitive hashing, global-local state to decrease the attention complexity to nearly linear. Our cross-block reader is based on ETC~\citep{ainslie2020etc}, but unlike prior works that process one long document for QA, our task requires reading multiple blocks containing both structured and unstructured data. To handle the long sequence of retrieved documents in open-domain question answering, Fusion-in-Decoder~\citep{izacard2020leveraging} has been proposed to replace the extractive model with an encoder-decoder generative model. The long sequence of passages are split and encoded independently to decrease the computation complexity, but the decoder still uses full attention over the tens of thousands of encoded vectors to generate the answer token by token. Such full-attention can decrease the decoding speed by an order of magnitude, while our sparse-attention-based cross-block reader can still maintain the same speed as the standard BERT model. 
\section{Conclusion}
We focus on the problem of performing open question answering over tables and text in this paper. By proposing the fusion retriever and sparse reader, we manage the increase the model's effectiveness and efficiency by a large margin. One interesting question we would like to ask in the future is: can we extend open question answering system to more modalities like images or audios, etc? 
%Some questions can be better answered by images, audios and other resources, but the task can be drastically more challenging by including more modalities, as we have learned from this paper.

\bibliography{fused}

\begin{thebibliography}{43}
\providecommand{\natexlab}[1]{#1}
\providecommand{\url}[1]{\texttt{#1}}
\expandafter\ifx\csname urlstyle\endcsname\relax
  \providecommand{\doi}[1]{doi: #1}\else
  \providecommand{\doi}{doi: \begingroup \urlstyle{rm}\Url}\fi

\bibitem[Abadi et~al.(2016)Abadi, Barham, Chen, Chen, Davis, Dean, Devin,
  Ghemawat, Irving, Isard, et~al.]{abadi2016tensorflow}
Mart{\'\i}n Abadi, Paul Barham, Jianmin Chen, Zhifeng Chen, Andy Davis, Jeffrey
  Dean, Matthieu Devin, Sanjay Ghemawat, Geoffrey Irving, Michael Isard, et~al.
\newblock Tensorflow: A system for large-scale machine learning.
\newblock In \emph{12th $\{$USENIX$\}$ symposium on operating systems design
  and implementation ($\{$OSDI$\}$ 16)}, pp.\  265--283, 2016.

\bibitem[Ainslie et~al.(2020)Ainslie, Ontanon, Alberti, Pham, Ravula, and
  Sanghai]{ainslie2020etc}
Joshua Ainslie, Santiago Ontanon, Chris Alberti, Philip Pham, Anirudh Ravula,
  and Sumit Sanghai.
\newblock Etc: Encoding long and structured data in transformers.
\newblock \emph{Proceedings of EMNLP 2020}, 2020.

\bibitem[Asai et~al.(2019)Asai, Hashimoto, Hajishirzi, Socher, and
  Xiong]{asai2019learning}
Akari Asai, Kazuma Hashimoto, Hannaneh Hajishirzi, Richard Socher, and Caiming
  Xiong.
\newblock Learning to retrieve reasoning paths over wikipedia graph for
  question answering.
\newblock In \emph{International Conference on Learning Representations}, 2019.

\bibitem[Beltagy et~al.(2020)Beltagy, Peters, and Cohan]{beltagy2020longformer}
Iz~Beltagy, Matthew~E Peters, and Arman Cohan.
\newblock Longformer: The long-document transformer.
\newblock \emph{arXiv preprint arXiv:2004.05150}, 2020.

\bibitem[Bromley et~al.(1994)Bromley, Guyon, LeCun, S{\"a}ckinger, and
  Shah]{bromley1994signature}
Jane Bromley, Isabelle Guyon, Yann LeCun, Eduard S{\"a}ckinger, and Roopak
  Shah.
\newblock Signature verification using a" siamese" time delay neural network.
\newblock In \emph{Advances in neural information processing systems}, pp.\
  737--744, 1994.

\bibitem[Chakrabarti et~al.(2020)Chakrabarti, Chen, Shakeri, and
  Cao]{chakrabarti2020open}
Kaushik Chakrabarti, Zhimin Chen, Siamak Shakeri, and Guihong Cao.
\newblock Open domain question answering using web tables.
\newblock \emph{arXiv preprint arXiv:2001.03272}, 2020.

\bibitem[Chen et~al.(2017)Chen, Fisch, Weston, and Bordes]{chen2017reading}
Danqi Chen, Adam Fisch, Jason Weston, and Antoine Bordes.
\newblock Reading wikipedia to answer open-domain questions.
\newblock In \emph{Proceedings of the 55th Annual Meeting of the Association
  for Computational Linguistics (Volume 1: Long Papers)}, pp.\  1870--1879,
  2017.

\bibitem[Chen et~al.(2019)Chen, Wang, Chen, Zhang, Wang, Li, Zhou, and
  Wang]{chen2019tabfact}
Wenhu Chen, Hongmin Wang, Jianshu Chen, Yunkai Zhang, Hong Wang, Shiyang Li,
  Xiyou Zhou, and William~Yang Wang.
\newblock Tabfact: A large-scale dataset for table-based fact verification.
\newblock In \emph{International Conference on Learning Representations}, 2019.

\bibitem[Chen et~al.(2020)Chen, Zha, Chen, Xiong, Wang, and
  Wang]{chen2020hybridqa}
Wenhu Chen, Hanwen Zha, Zhiyu Chen, Wenhan Xiong, Hong Wang, and William Wang.
\newblock Hybridqa: A dataset of multi-hop question answering over tabular and
  textual data.
\newblock \emph{Proceedings of Findings of EMNLP 2020}, 2020.

\bibitem[Child et~al.(2019)Child, Gray, Radford, and
  Sutskever]{child2019generating}
Rewon Child, Scott Gray, Alec Radford, and Ilya Sutskever.
\newblock Generating long sequences with sparse transformers.
\newblock \emph{arXiv preprint arXiv:1904.10509}, 2019.

\bibitem[De~Cao et~al.(2020)De~Cao, Izacard, Riedel, and
  Petroni]{de2020autoregressive}
Nicola De~Cao, Gautier Izacard, Sebastian Riedel, and Fabio Petroni.
\newblock Autoregressive entity retrieval.
\newblock \emph{arXiv preprint arXiv:2010.00904}, 2020.

\bibitem[Devlin et~al.(2019)Devlin, Chang, Lee, and Toutanova]{devlin2019bert}
Jacob Devlin, Ming-Wei Chang, Kenton Lee, and Kristina Toutanova.
\newblock Bert: Pre-training of deep bidirectional transformers for language
  understanding.
\newblock In \emph{Proceedings of the 2019 Conference of the North American
  Chapter of the Association for Computational Linguistics: Human Language
  Technologies, Volume 1 (Long and Short Papers)}, pp.\  4171--4186, 2019.

\bibitem[Dhingra et~al.(2019)Dhingra, Zaheer, Balachandran, Neubig,
  Salakhutdinov, and Cohen]{dhingra2019differentiable}
Bhuwan Dhingra, Manzil Zaheer, Vidhisha Balachandran, Graham Neubig, Ruslan
  Salakhutdinov, and William~W Cohen.
\newblock Differentiable reasoning over a virtual knowledge base.
\newblock In \emph{International Conference on Learning Representations}, 2019.

\bibitem[Ding et~al.(2019)Ding, Zhou, Chen, Yang, and Tang]{ding2019cognitive}
Ming Ding, Chang Zhou, Qibin Chen, Hongxia Yang, and Jie Tang.
\newblock Cognitive graph for multi-hop reading comprehension at scale.
\newblock In \emph{Proceedings of the 57th Annual Meeting of the Association
  for Computational Linguistics}, pp.\  2694--2703, 2019.

\bibitem[Guu et~al.(2020)Guu, Lee, Tung, Pasupat, and Chang]{guu2020realm}
Kelvin Guu, Kenton Lee, Zora Tung, Panupong Pasupat, and Ming-Wei Chang.
\newblock Realm: Retrieval-augmented language model pre-training.
\newblock \emph{Proceedings of ICML 2020}, 2020.

\bibitem[Herzig et~al.(2020)Herzig, Nowak, M{\"u}ller, Piccinno, and
  Eisenschlos]{herzig2020tapas}
Jonathan Herzig, Pawe{\l}~Krzysztof Nowak, Thomas M{\"u}ller, Francesco
  Piccinno, and Julian~Martin Eisenschlos.
\newblock Tapas: Weakly supervised table parsing via pre-training.
\newblock \emph{ACL 2020}, 2020.

\bibitem[Izacard \& Grave(2020)Izacard and Grave]{izacard2020leveraging}
Gautier Izacard and Edouard Grave.
\newblock Leveraging passage retrieval with generative models for open domain
  question answering.
\newblock \emph{arXiv preprint arXiv:2007.01282}, 2020.

\bibitem[Joshi et~al.(2017)Joshi, Choi, Weld, and
  Zettlemoyer]{joshi2017triviaqa}
Mandar Joshi, Eunsol Choi, Daniel~S Weld, and Luke Zettlemoyer.
\newblock Triviaqa: A large scale distantly supervised challenge dataset for
  reading comprehension.
\newblock In \emph{Proceedings of the 55th Annual Meeting of the Association
  for Computational Linguistics (Volume 1: Long Papers)}, pp.\  1601--1611,
  2017.

\bibitem[Karpukhin et~al.(2020)Karpukhin, O{\u{g}}uz, Min, Wu, Edunov, Chen,
  and Yih]{karpukhin2020dense}
Vladimir Karpukhin, Barlas O{\u{g}}uz, Sewon Min, Ledell Wu, Sergey Edunov,
  Danqi Chen, and Wen-tau Yih.
\newblock Dense passage retrieval for open-domain question answering.
\newblock \emph{EMNLP 2020}, 2020.

\bibitem[Kitaev et~al.(2020)Kitaev, Kaiser, and Levskaya]{kitaev2020reformer}
Nikita Kitaev, {\L}ukasz Kaiser, and Anselm Levskaya.
\newblock Reformer: The efficient transformer.
\newblock \emph{ICLR}, 2020.

\bibitem[Kwiatkowski et~al.(2019)Kwiatkowski, Palomaki, Redfield, Collins,
  Parikh, Alberti, Epstein, Polosukhin, Devlin, Lee,
  et~al.]{kwiatkowski2019natural}
Tom Kwiatkowski, Jennimaria Palomaki, Olivia Redfield, Michael Collins, Ankur
  Parikh, Chris Alberti, Danielle Epstein, Illia Polosukhin, Jacob Devlin,
  Kenton Lee, et~al.
\newblock Natural questions: a benchmark for question answering research.
\newblock \emph{Transactions of the Association for Computational Linguistics},
  7:\penalty0 453--466, 2019.

\bibitem[Lee et~al.(2019)Lee, Chang, and Toutanova]{lee2019latent}
Kenton Lee, Ming-Wei Chang, and Kristina Toutanova.
\newblock Latent retrieval for weakly supervised open domain question
  answering.
\newblock In \emph{Proceedings of the 57th Annual Meeting of the Association
  for Computational Linguistics}, pp.\  6086--6096, 2019.

\bibitem[Liu et~al.(2019)Liu, Ott, Goyal, Du, Joshi, Chen, Levy, Lewis,
  Zettlemoyer, and Stoyanov]{liu2019roberta}
Yinhan Liu, Myle Ott, Naman Goyal, Jingfei Du, Mandar Joshi, Danqi Chen, Omer
  Levy, Mike Lewis, Luke Zettlemoyer, and Veselin Stoyanov.
\newblock Roberta: A robustly optimized bert pretraining approach.
\newblock \emph{arXiv preprint arXiv:1907.11692}, 2019.

\bibitem[Loshchilov \& Hutter(2019)Loshchilov and
  Hutter]{loshchilov2017decoupled}
Ilya Loshchilov and Frank Hutter.
\newblock Decoupled weight decay regularization.
\newblock \emph{ICLR 2019}, 2019.

\bibitem[Min et~al.(2019)Min, Chen, Zettlemoyer, and
  Hajishirzi]{min2019knowledge}
Sewon Min, Danqi Chen, Luke Zettlemoyer, and Hannaneh Hajishirzi.
\newblock Knowledge guided text retrieval and reading for open domain question
  answering.
\newblock \emph{arXiv preprint arXiv:1911.03868}, 2019.

\bibitem[Nie et~al.(2019)Nie, Wang, and Bansal]{nie2019revealing}
Yixin Nie, Songhe Wang, and Mohit Bansal.
\newblock Revealing the importance of semantic retrieval for machine reading at
  scale.
\newblock In \emph{Proceedings of the 2019 Conference on Empirical Methods in
  Natural Language Processing and the 9th International Joint Conference on
  Natural Language Processing (EMNLP-IJCNLP)}, pp.\  2553--2566, 2019.

\bibitem[Parikh et~al.(2020)Parikh, Wang, Gehrmann, Faruqui, Dhingra, Yang, and
  Das]{parikh2020totto}
Ankur~P Parikh, Xuezhi Wang, Sebastian Gehrmann, Manaal Faruqui, Bhuwan
  Dhingra, Diyi Yang, and Dipanjan Das.
\newblock Totto: A controlled table-to-text generation dataset.
\newblock \emph{arXiv preprint arXiv:2004.14373}, 2020.

\bibitem[Pasupat \& Liang(2015)Pasupat and Liang]{pasupat2015compositional}
Panupong Pasupat and Percy Liang.
\newblock Compositional semantic parsing on semi-structured tables.
\newblock In \emph{Proceedings of the 53rd Annual Meeting of the Association
  for Computational Linguistics and the 7th International Joint Conference on
  Natural Language Processing (Volume 1: Long Papers)}, pp.\  1470--1480, 2015.

\bibitem[Petroni et~al.(2020)Petroni, Piktus, Fan, Lewis, Yazdani, De~Cao,
  Thorne, Jernite, Plachouras, Rockt{\"a}schel, et~al.]{petroni2020kilt}
Fabio Petroni, Aleksandra Piktus, Angela Fan, Patrick Lewis, Majid Yazdani,
  Nicola De~Cao, James Thorne, Yacine Jernite, Vassilis Plachouras, Tim
  Rockt{\"a}schel, et~al.
\newblock Kilt: a benchmark for knowledge intensive language tasks.
\newblock \emph{arXiv preprint arXiv:2009.02252}, 2020.

\bibitem[Qi et~al.(2019)Qi, Lin, Mehr, Wang, and Manning]{qi2019answering}
Peng Qi, Xiaowen Lin, Leo Mehr, Zijian Wang, and Christopher~D Manning.
\newblock Answering complex open-domain questions through iterative query
  generation.
\newblock In \emph{Proceedings of the 2019 Conference on Empirical Methods in
  Natural Language Processing and the 9th International Joint Conference on
  Natural Language Processing (EMNLP-IJCNLP)}, pp.\  2590--2602, 2019.

\bibitem[Radford et~al.(2019)Radford, Wu, Child, Luan, Amodei, and
  Sutskever]{radford2019language}
Alec Radford, Jeffrey Wu, Rewon Child, David Luan, Dario Amodei, and Ilya
  Sutskever.
\newblock Language models are unsupervised multitask learners.
\newblock \emph{OpenAI Blog}, 1\penalty0 (8):\penalty0 9, 2019.

\bibitem[Robertson \& Zaragoza(2009)Robertson and
  Zaragoza]{robertson2009probabilistic}
Stephen Robertson and Hugo Zaragoza.
\newblock \emph{The probabilistic relevance framework: BM25 and beyond}.
\newblock Now Publishers Inc, 2009.

\bibitem[Roy et~al.(2020)Roy, Saffar, Vaswani, and Grangier]{roy2020efficient}
Aurko Roy, Mohammad Saffar, Ashish Vaswani, and David Grangier.
\newblock Efficient content-based sparse attention with routing transformers.
\newblock \emph{arXiv preprint arXiv:2003.05997}, 2020.

\bibitem[Sun et~al.(2018)Sun, Dhingra, Zaheer, Mazaitis, Salakhutdinov, and
  Cohen]{sun-etal-2018-open}
Haitian Sun, Bhuwan Dhingra, Manzil Zaheer, Kathryn Mazaitis, Ruslan
  Salakhutdinov, and William Cohen.
\newblock Open domain question answering using early fusion of knowledge bases
  and text.
\newblock In \emph{Proceedings of the 2018 Conference on Empirical Methods in
  Natural Language Processing}, pp.\  4231--4242, Brussels, Belgium,
  October-November 2018. Association for Computational Linguistics.
\newblock \doi{10.18653/v1/D18-1455}.
\newblock URL \url{https://www.aclweb.org/anthology/D18-1455}.

\bibitem[Sun et~al.(2019)Sun, Bedrax-Weiss, and Cohen]{sun-etal-2019-pullnet}
Haitian Sun, Tania Bedrax-Weiss, and William Cohen.
\newblock {P}ull{N}et: Open domain question answering with iterative retrieval
  on knowledge bases and text.
\newblock In \emph{Proceedings of the 2019 Conference on Empirical Methods in
  Natural Language Processing and the 9th International Joint Conference on
  Natural Language Processing (EMNLP-IJCNLP)}, pp.\  2380--2390, Hong Kong,
  China, November 2019. Association for Computational Linguistics.
\newblock \doi{10.18653/v1/D19-1242}.
\newblock URL \url{https://www.aclweb.org/anthology/D19-1242}.

\bibitem[Sun et~al.(2016)Sun, Ma, He, Yih, Su, and Yan]{sun2016table}
Huan Sun, Hao Ma, Xiaodong He, Wen-tau Yih, Yu~Su, and Xifeng Yan.
\newblock Table cell search for question answering.
\newblock In \emph{Proceedings of the 25th International Conference on World
  Wide Web}, pp.\  771--782, 2016.

\bibitem[Yang et~al.(2018)Yang, Qi, Zhang, Bengio, Cohen, Salakhutdinov, and
  Manning]{yang2018hotpotqa}
Zhilin Yang, Peng Qi, Saizheng Zhang, Yoshua Bengio, William Cohen, Ruslan
  Salakhutdinov, and Christopher~D Manning.
\newblock Hotpotqa: A dataset for diverse, explainable multi-hop question
  answering.
\newblock In \emph{Proceedings of the 2018 Conference on Empirical Methods in
  Natural Language Processing}, pp.\  2369--2380, 2018.

\bibitem[Yih et~al.(2011)Yih, Toutanova, Platt, and Meek]{yih2011learning}
Wen-tau Yih, Kristina Toutanova, John~C Platt, and Christopher Meek.
\newblock Learning discriminative projections for text similarity measures.
\newblock In \emph{Proceedings of the fifteenth conference on computational
  natural language learning}, pp.\  247--256, 2011.

\bibitem[Yin et~al.(2020)Yin, Neubig, Yih, and Riedel]{yin2020tabert}
Pengcheng Yin, Graham Neubig, Wen-tau Yih, and Sebastian Riedel.
\newblock Tabert: Pretraining for joint understanding of textual and tabular
  data.
\newblock \emph{ACL 2020}, 2020.

\bibitem[Yu et~al.(2018)Yu, Zhang, Yang, Yasunaga, Wang, Li, Ma, Li, Yao,
  Roman, et~al.]{yu2018spider}
Tao Yu, Rui Zhang, Kai Yang, Michihiro Yasunaga, Dongxu Wang, Zifan Li, James
  Ma, Irene Li, Qingning Yao, Shanelle Roman, et~al.
\newblock Spider: A large-scale human-labeled dataset for complex and
  cross-domain semantic parsing and text-to-sql task.
\newblock In \emph{Proceedings of the 2018 Conference on Empirical Methods in
  Natural Language Processing}, pp.\  3911--3921, 2018.

\bibitem[Zaheer et~al.(2020)Zaheer, Guruganesh, Dubey, Ainslie, Alberti,
  Ontanon, Pham, Ravula, Wang, Yang, et~al.]{zaheer2020big}
Manzil Zaheer, Guru Guruganesh, Avinava Dubey, Joshua Ainslie, Chris Alberti,
  Santiago Ontanon, Philip Pham, Anirudh Ravula, Qifan Wang, Li~Yang, et~al.
\newblock Big bird: Transformers for longer sequences.
\newblock \emph{arXiv preprint arXiv:2007.14062}, 2020.

\bibitem[Zhao et~al.(2019)Zhao, Xiong, Rosset, Song, Bennett, and
  Tiwary]{zhao2019transformer}
Chen Zhao, Chenyan Xiong, Corby Rosset, Xia Song, Paul Bennett, and Saurabh
  Tiwary.
\newblock Transformer-xh: Multi-evidence reasoning with extra hop attention.
\newblock In \emph{International Conference on Learning Representations}, 2019.

\bibitem[Zhong et~al.(2017)Zhong, Xiong, and Socher]{zhong2017seq2sql}
Victor Zhong, Caiming Xiong, and Richard Socher.
\newblock Seq2sql: Generating structured queries from natural language using
  reinforcement learning.
\newblock \emph{arXiv preprint arXiv:1709.00103}, 2017.

\end{thebibliography}
\bibliographystyle{iclr2021_conference}

\clearpage
\appendix
\section{Dataset Collection}

\subsection{Dataset Annotation}
\label{sec:dataset_annotation}

\paragraph{Filtering}
The original HybridQA dataset contains over 72$k$ questions paired with 13$k$ hyperlinked tables. We adopt two filtering heuristics to make the decontextualization easier.  First, we filter out tables without enough meta-information or containing too much non-textual information\footnote{Like longitude, latitude, mathematical formulae, etc}. Second, we filter out overly-long questions, i.e., questions longer than 30 words. These two filtering heuristics result in a cleaner subset of 46$k$ questions paired with 9$k$ in-domain tables.

\paragraph{Quality Control}
\label{sec:quality_control}
During annotation, we conduct strict manual quality evaluation over the de-contextualized questions, with the following criteria: 1) the annotated question retains the same semantics and answer as before, 2) the annotated question still requires multi-hop reasoning over both table and passages, and 3) the annotated question is concise and fluent. The manual quality checking was performed over batches distributed to the same annotator. Each batch consists of six questions, one of which will be sampled to decide the acceptance/rejection of the whole batch. The overall acceptance rate for the crowd-sourcing job is 71\%, and a rejected job was re-distributed until it was accepted.  

\subsection{Dataset Examples}
\begin{figure}[!thb]
    \centering
    \includegraphics[width=0.98\linewidth]{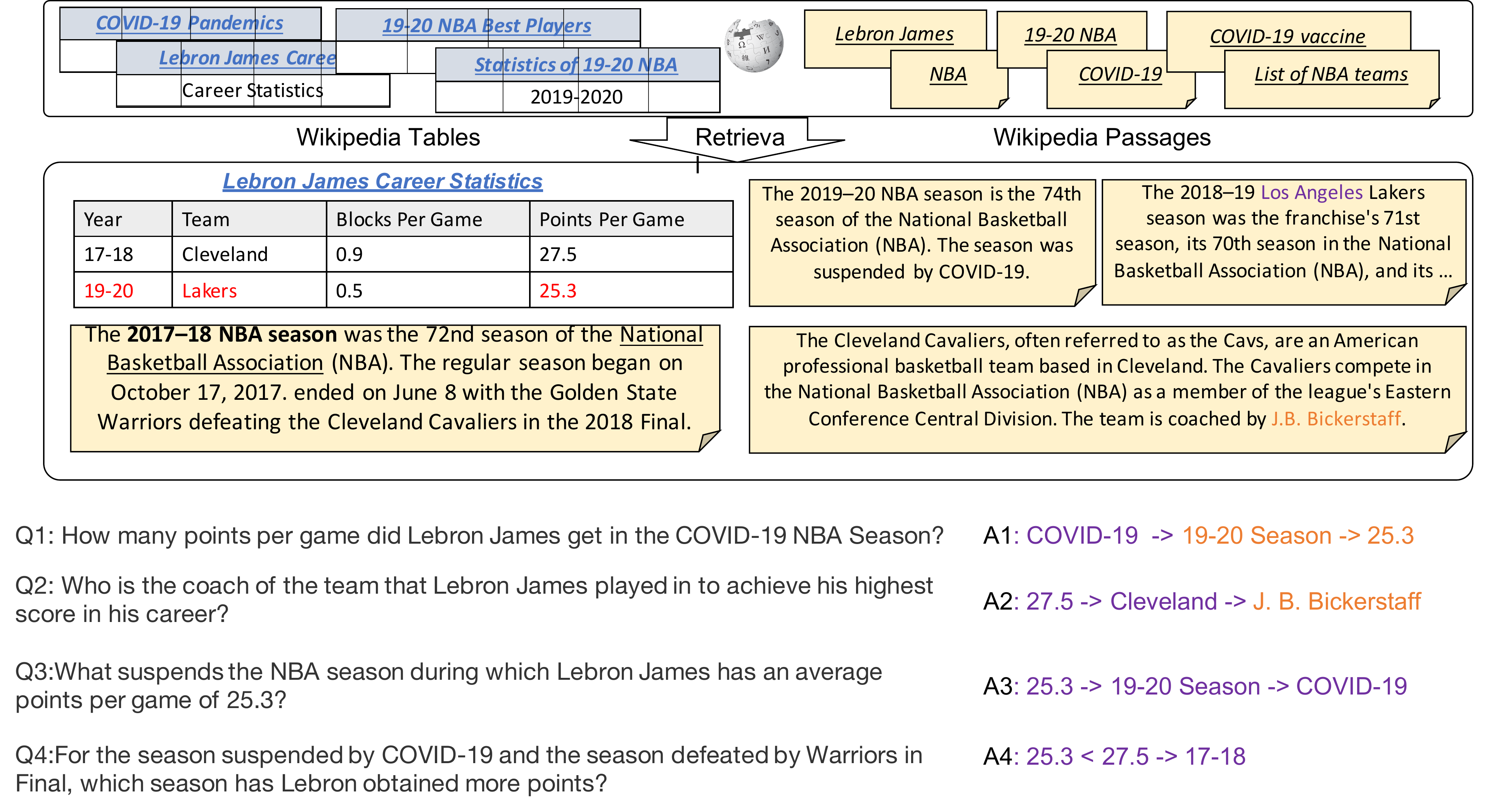}
    \vspace{-2ex}
    \caption{More examples from \dataname}
    \label{fig:more_examples}
\end{figure}

We demonstrate more examples in~\autoref{fig:more_examples}, which includes more diverse inference chains, like table $\rightarrow$ text; text  $\rightarrow$ table  $\rightarrow$ text; text + text  $\rightarrow$ comparative  $\rightarrow$ table. Our model is able to perform these reasoning types quite well by jointly matching a query against a fused table-text block.

\subsection{Question Types}
\label{section:question_type}
We randomly sampled 100 questions from the dataset to manually analyze the kinds of inference chains seen in \dataname and divide the major types into the following categories:
\begin{enumerate}
    \item Single hop questions (13\%) require reading one table or one passage to answer.
    \item Two hop questions (57\%) require reading one passage and one table to answer.  These can be subclassified as `table bridge'  $\rightarrow$ `answer text'\footnote{I.e., a table forms a bridge between a question and a textual passage, which is read in the first hop, and the answer is extracted from the text.} or `text bridge' $\rightarrow$ `answer table'.
    \item Multi-hop questions (30\%) require reading two passages and one table to answer. These mainly following the reasoning chain of `text bridge' $\rightarrow$ `table bridge'  $\rightarrow$ `answer text'. 
    \item Questions with multiple reasoning paths: Due to information redundancy in Wikipedia, similar information can appear in both tables and text. We find that 9\% of questions are answerable by reading one text passage, 18\% of questions are answerable by reading two text passages and 4\% of questions are answerable by reading two tables.
\end{enumerate}

\section{Model Details}
\subsection{Retrieval Block Representation}
\label{section:table_seg_representation}
\begin{figure}[!thb]
    \centering
    \includegraphics[width=0.9\linewidth]{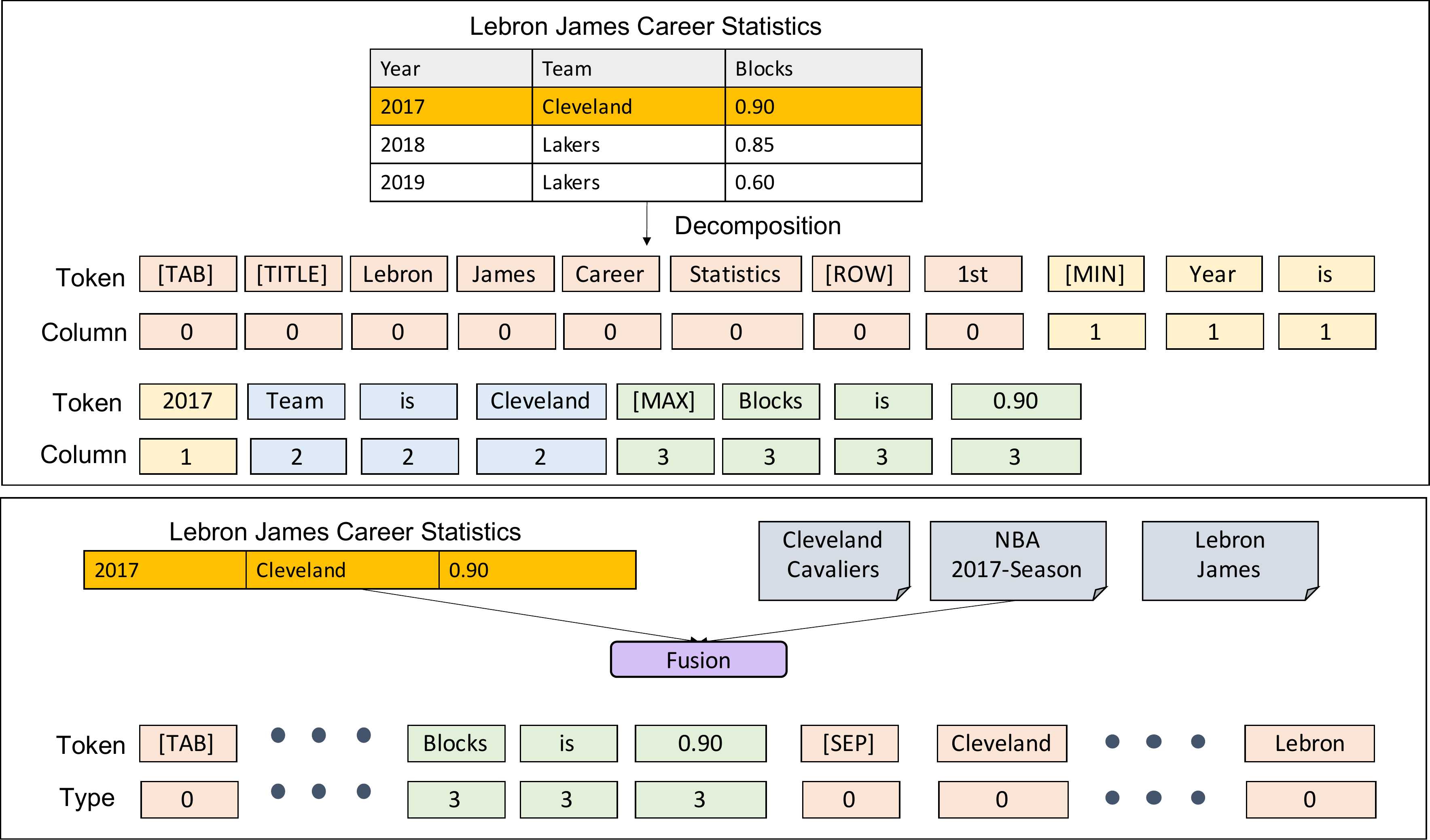}
    \caption{The decomposition of the original table into segments. }
    \label{fig:decompose}
\end{figure}

The table decomposition is visualized in~\autoref{fig:decompose}. The title/section title are prefixed to the table segment. We add the row position token `1st' and a max/min special token over the column to infuse global table information into the segmented unit. The column embedding is added as another vector to the representation. The table segment representation is relatively small and easy to deal with in the following reader model.
After the table-passage alignment, we group the highly related units together and represent them as the lower part demonstrated in~\autoref{fig:decompose}. We add [SEP] tokens to separate different passages and set their type id to 0. Such a flattened representation for fused block $b_F$ will be used throughout our experiments for both sparse/dense retriever and ETC reader.

\subsection{Iterative Retriever}
\label{section:iterative_retriever}
Iterative retrieval has been used in recent graph-based multi-hop retrieval models to gradually retrieve documents to find the correct supporting evidence. Specifically, the retriever conditions the $i$-th round retrieval on the previous round of retrieval results.

\paragraph{Sparse Retriever}
The sparse retriever uses uni-gram lexical feature to compute the BM-25 score between the $q, .., b_{1..j-1}$ over $b_j \in \mathbb{B}$ to get the top candidates. Here we describe a two-step retrieval procedure, called here the LxM procedure. In the first step, the model calculates the BM25 score between question over all the candidates in $\mathbb{B}$ to select top L/2 table segments $b_T$, and L/2 passages $b_P$. In the second step, the question is concatenated with the retrieved table segment to form L/2 new queries $[q;b_T]$ which are used to retrieve LM/2 passages from $\mathbb{B}$. The question is also concatenated with the retrieved passage titles to form another K/2 queries $[q;b_P]$ to retrieve LM/2 table segments. The retrieval procedure results in at most LM+L unique blocks. Each unique block aggregates its score from two rounds, denoted as $f(q,b)$, which is used to rank the top-K candidates for the next step. We truncate the top-K candidate by thresholding their combined length.

\paragraph{Dual-Encoder Retriever}
The dual encoder uses a BERT-based encoder to compress each question, table segment, and passage into a fixed-length vector and then computes the dot product between fixed vectors to obtain the highest scored candidates from pool $\mathbb{B}$. However, since the dataset does not provide an explicit supervision signal for the iterative retrieval, we heuristically synthesize some noisy retrieval chains using lexical matching. The retrieval inference chain is depicted as $b_1 \rightarrow b_2 \rightarrow b_K$, which is used to train the model $f(b_k|q, b_{1..k-1})$ in a supervised manner. At inference time, the dual encoder retriever will encode a query $q$ into a fixed vector and retrieve the first $L$ blocks from $\mathbb{B}$. The blocks are appended to query $q$ to form $L$ new queries $[q;b_i]$, which is re-encoded and search for $LM$ new neighbors. We experiment with a maximum of 3-step retrieval of LxMxN to obtain a maximum of L+LxM+LMN unique blocks. Similarly, each unique block aggregates its score from different rounds to select the top-K candidates for the next step.

\subsection{Sparse Fused Retriever}
The sparse fused retriever uses the uni-gram lexical feature to compute the BM-25 score between $q$ over $b_F \in \mathbb{B}_F$. The uni-gram feature of $b_F$ is based on the representation depicted in~\autoref{section:table_seg_representation}. Note that this BM25 feature will be much more abundant than the BM25 feature in iterative sparse retriever because it encloses more uni-grams. Instead of doing multiple rounds of retrieval, the fused retrieval once retrieve once over the candidate pool and treat all the units inside the block as the same retrieval score. Finally, We truncate the top-K candidate by thresholding their combined length.

\subsection{Query Augmentation}
\begin{figure}[!thb]
    \centering
    \includegraphics[width=0.88\linewidth]{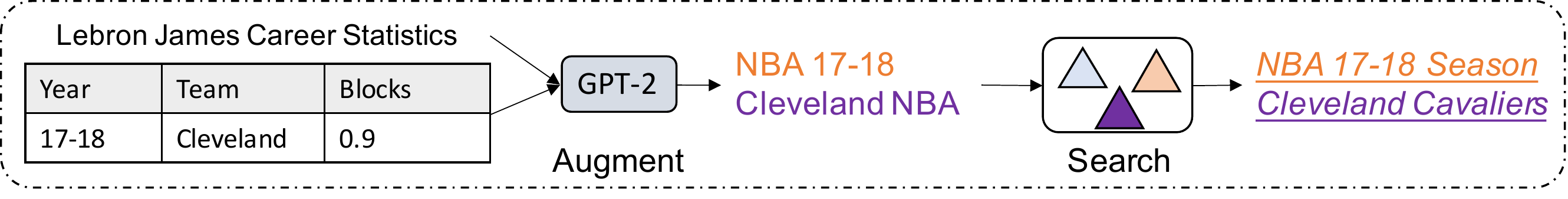}
    \vspace{-2ex}
    \caption{Fusion: 1) GPT-2 query augmentation, 2) nearest neighbor search over passages.}
    \label{fig:linker}
\end{figure}

The query augmentation procedure is depicted in~\autoref{fig:linker}.

\subsection{Dense Retrieval/In-Batch Negative}
\label{section:in_batch_negative}
Recently, different dense-retrieval methods~\citep{lee2019latent,guu2020realm,karpukhin2020dense} based on dual-encoders~\citep{bromley1994signature} have been shown to surpass traditional sparse retrieval in open-QA models.
The query and the passage are both encoded using a Transformer, which produces a vector for every token. As in \citep{devlin2019bert}, the vector corresponding to the first token, $\cls$, which is used as a ``pooled'' representation of the sequence (denoted $\bertsub{CLS}$). The dense retrieval function can be represented as the dot product between $\bertsub{CLS}(q)$ and $\bertsub{CLS}(p)$ for each document in the text collection, much like TF-IDF~\citep{chen2017reading} and BM25~\citep{robertson2009probabilistic} on some Open QA datasets. To train the dual-encoder, the in-batch negative trick~\citep{yih2011learning,karpukhin2020dense} plays an important role, which uses B training instances in each batch and views the other B-1 instances inside the batch as the negatives. In this way, the model reuses computation and effectively trains on $B^2$ question/document pairs in each batch.

\section{Performance Analysis}
\label{section:error_analysis}
\subsection{Question Type Breakdown performance}
We measure our best model's performance (dense fusion retriever + cross-block reader) and baseline model (dense Iterative-Retriever + single-block reader) on different question types~
(\autoref{section:question_type}) to show the breakdown statistics in~\autoref{fig:breakdown1} and ~\autoref{fig:breakdown}. As we can observe, the gap between our model vs. baseline in 1-hop question is less significant as 2-hop and 3-hop questions. The iterative retriever's performance is sensitive to the number of hops in the question, which is the largely due to the error propagation in the beam search stage. If the retriever fails to include the golden block in the earlier stage beam, the retrieval in later stage cannot recover from such failure. In contrast, our fusion retriever can group the related information prior to retrieval to retrieve all the blocks at once, which makes the model less prone to the error propagation issue. Another reason is due to the cross-block reader, which can reason over different blocks in the latent space, such implicit reasoning can also decrease the error propagation issue. To sum up, our model is more powerful to deal with complex multi-hop open questions with much less performance drop. 
\begin{figure}[!htb]
\centering
\begin{minipage}[t]{.45\textwidth}
  \centering
    \begin{tikzpicture}
\begin{axis}[
    height=1.2in, 
    width=0.88\linewidth,
    scale only axis,
    ybar=3pt,
    enlargelimits=0.15,
    legend style={
     at={(0.05,0.95)},
      anchor=north west,
      legend columns=-1
      },
    ymin=5,
    ymax=50,
    xticklabel style = {font=\small,yshift=0.5ex},
    symbolic x coords={
    1-hop,
    2-hop,
    3-hop,
    },
    xtick=data,
    nodes near coords,
    nodes near coords align={vertical},
    every node near coord/.append style={font=\tiny},
    ]
\addplot[orange!20!black,fill=orange!70!white] coordinates {
(1-hop, 23.0) (2-hop, 8.7) (3-hop, 3.3)
};
\addplot[blue!20!black,fill=blue!70!white]  coordinates {
(1-hop, 26.6) (2-hop, 11.9) (3-hop, 5.7)
};

\legend{EM, F1}
\end{axis}
\end{tikzpicture}
  \caption{Breakdown for iterative retriever + sing-block reader.}
  \label{fig:breakdown1}
\end{minipage}
\begin{minipage}[t]{.45\textwidth}
  \centering
    \begin{tikzpicture}
\begin{axis}[
    height=1.2in, 
    width=0.88\linewidth,
    scale only axis,
    ybar=3pt,
    enlargelimits=0.15,
    legend style={
     at={(0.05,0.95)},
      anchor=north west,
      legend columns=-1
      },
    % ylabel={Metric \%},\
    ymin=5,
    ymax=50,
    xticklabel style = {font=\small,yshift=0.5ex},
    symbolic x coords={
    1-hop,
    2-hop,
    3-hop,
    },
    xtick=data,
    nodes near coords,
    nodes near coords align={vertical},
    every node near coord/.append style={font=\tiny},
    ]
\addplot[orange!20!black,fill=orange!70!white] coordinates {
(1-hop, 30.7) (2-hop, 24.5) (3-hop, 20.0)
};
\addplot[blue!20!black,fill=blue!70!white]  coordinates {
(1-hop, 34.3) (2-hop, 28.1) (3-hop, 23.4)
};

\legend{EM, F1}
\end{axis}
\end{tikzpicture}
  \caption{Breakdown for fusion retriever + cross-block reader.}
  \label{fig:breakdown}
\end{minipage}%
\end{figure}

\subsection{Retriever Error Analysis}
\begin{figure}[!thb]
    \centering
    \includegraphics[width=0.9\linewidth]{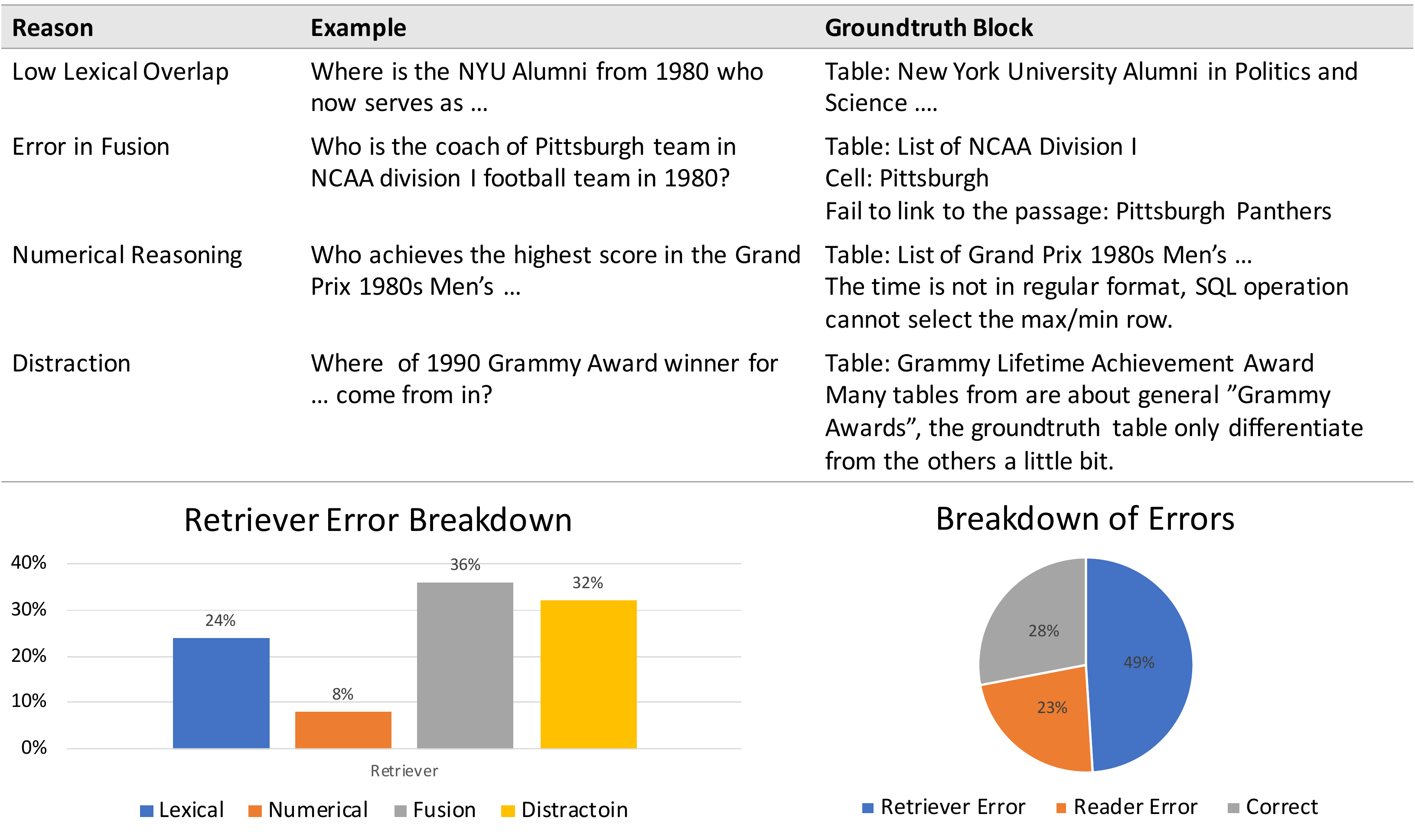}
    \caption{The main error types in the retriever. }
    \label{fig:error_analysis}
\end{figure}

We conduct error analysis to see what are the major issues with the retriever and conclude the following types in~\autoref{fig:error_analysis}. The major issues causing the system to retrieve unrelated evidence are low lexical overlap, fusion errors, numerical reasoning and distracting passages or tables.
In the Low-Lexical-Overlap case, the errors are mainly coming from the abbreviation, rephrasing of the table metadata, for example, `New York University' is shortened as `NYU', etc. In the Fusion-Error case, the issue is mainly because the entity-linking model fails to fuse all the hyperlinked passages, the error (F1=50\%) is quantitatively reflected in the entity-linker-performance figure. Numerical-Reasoning error is mainly related to the failure to find max/min/earliest/latest row in the table. The distraction error is mainly caused by some distracting passages or tables having very similar information. We sample 50 error samples from the dev-set and attribute their errors to the above categories. As shown in the left part, we found that the numerical reasoning error is not as severe as the other three types because the proportion of questions requiring it is relatively small. Besides the low-lexical overlap error, which is general across other open QA datasets like NQ and HoptpotQA, we found the fusion and distraction errors quite specific in our dataset.
\begin{itemize}
    \item (Fusion) Questions which ask about tables which are linked to too many linked passages. For example, a question over table “Team Record” in \url{https://en.wikipedia.org/wiki/Sevens_Grand_Prix_Series} is hard because some table rows associate with over 10 passages, it’s hard to link them and fuse all of them into a fused block. 
    \item (Distraction) Questions which ask about topics which are contained by too many similar tables, it’s hard to differentiate the true one. For example,  there are over ten tables in \url{https://en.wikipedia.org/wiki/List_of_RMIT_University_people}, these similar tables can easily distract the attention of the retriever to select the wrong one from the same page.
\end{itemize}

From our quantitative results, we can attribute the errors to retriever and reader, among all the examples, 49\% of examples cannot find the correct supporting block. For the rest 51\% examples with correct block retrieved, the reader fails to select the correct span for 23\% of them.

\subsection{Length Sensitivity Analysis of Retrieval/Reader}
\begin{figure}[!h]
  \begin{center}
  \includegraphics[width=0.8\linewidth]{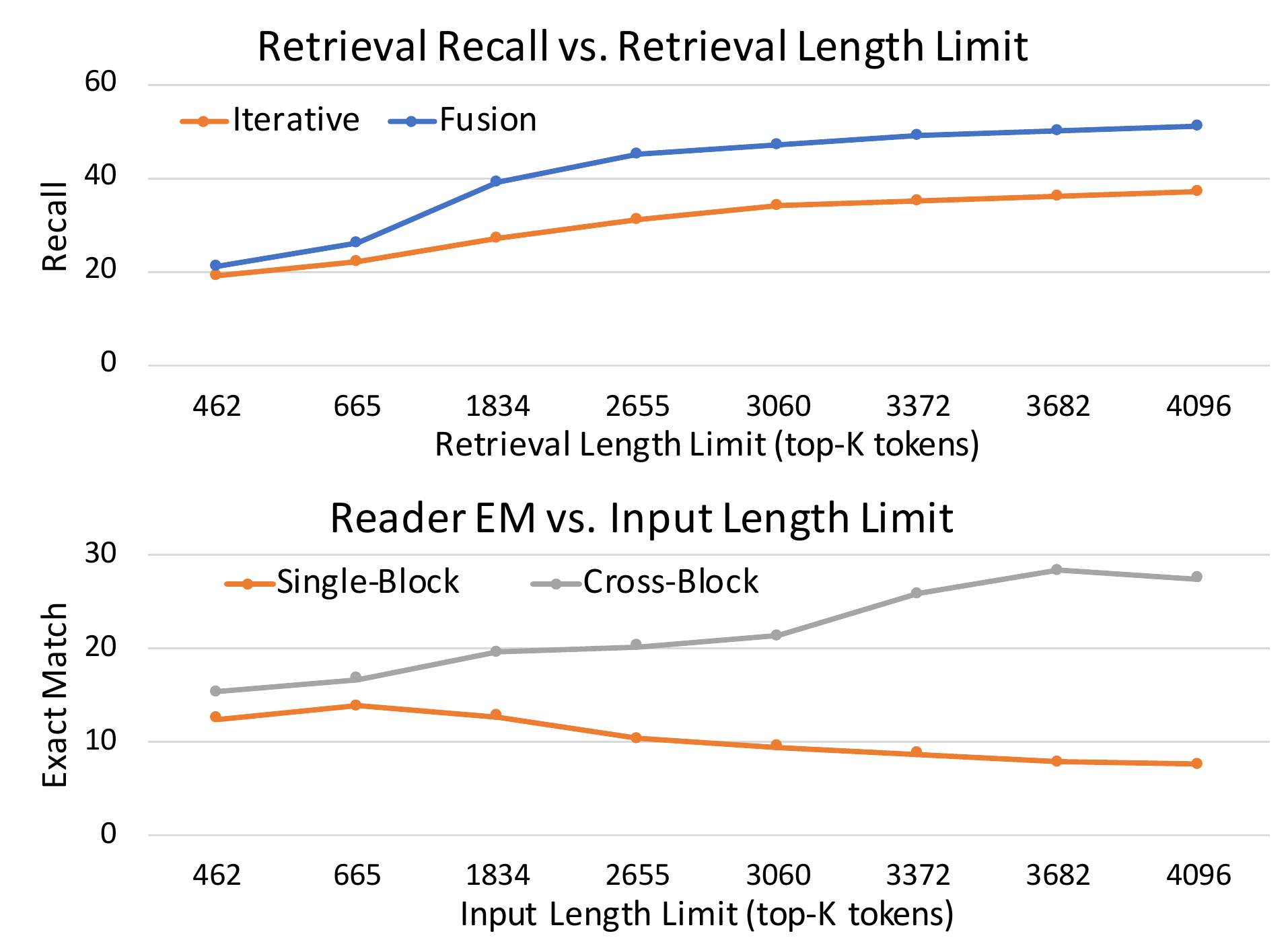}
\caption{Analyzing retriever performance. }
\label{fig:length_limit}
\end{center}
\vspace{-.15in}
\end{figure}

We perform sensitivity analysis for both retriever and reader in~\autoref{fig:length_limit}.
We gradually increase the length limit of retrieved evidence from 400 to 4096 to first visualize its impact on the sparse iterative and dense fusion retriever. For both fusion and iterative retriever, we can observe that both of their recall@K significantly improves as the length limit increases. With a low budget of token limit, their performance is gap is smaller because its performance is dominated by the single-hop questions in the dataset. As the length limit increases, the improvement for fusion retriever is steeper than iterative retriever because the contextualized fusion block becomes easier to retrieve than standalone table segment or passage.

We also visualize the input length's impact on the single-block The performance of single vs cross-block reader. With a low budget of token limit, both single-block and cross-block readers are comparable. However, as the limit increases to 4000, the cross-block reader can digest long input with its sparse attention mechanism to achieve better scores, while the single-block reader needs to truncate the information to read independently, which leads to a even lower EM score due to introduced noise. This observation reveals the importance of modeling cross-attention between different retrieved evidence units to reach a consistent answer. In single-block reader, dealing with different blocks independently can lead to suboptimal prediction in our dataset.

\section{Connection to Existing Work}
\paragraph{KB and Text}
The problem combining structured and unstructured data has been studied in question answering. The previous approaches are mainly divided into two categories: 1) FusionNet and PullNet~\citep{sun-etal-2018-open,sun-etal-2019-pullnet} simulate a KB-incomplete setting by masking out some triples from a knowledge graph and use textual information to complete the masked KB triples; these experiments are conducted on KB-based QA datasets. 2) DrKIT~\citep{dhingra2019differentiable} and Knowledge-Guided Retrieval~\citep{min2019knowledge} propose to use entity mentions and relations to guide the retrieval from the web. However, the KB is mainly used as an assisting tool, rather than a necessary information source. In \dataname, the structured data is used as necessary information in a realistic setting. The two information forms are combined in a non-trivial way, which makes the problem much harder than the other structure-unstructured QA settings.
\paragraph{Entity Linking}
Our generative entity linker is related to knowledge-enhanced language understanding~\citep{petroni2020kilt}, which proposes a seq2seq model to deal with different knowledge-intensive tasks like slot filling, entity linking, etc. There is a concurrent related work on auto-regressive entity linking~\citep{de2020autoregressive}, which also demonstrates the advantages of using an autoregressive generation model for entity retrieval. \vspace{1ex}\\

\end{document}